\documentclass[10pt,twocolumn]{article}
\usepackage[top=1.9cm, bottom=1.9cm, left=1.7cm, right=1.7cm]{geometry}
\usepackage{times}
\usepackage[hyphens]{url}  %
\usepackage[keeplastbox]{flushend}  %
\usepackage{graphicx}  %
\usepackage{xcolor}
\usepackage{flushend}
\usepackage{booktabs}
\usepackage{graphicx}
\usepackage{paralist}
\usepackage{enumitem}
\usepackage[small,bf]{caption}
\usepackage{subfigure}
\usepackage[hang,flushmargin]{footmisc}

\usepackage{amsmath}

\usepackage[bookmarks=true,%
            bookmarksnumbered=true,%
            colorlinks=true,%
            linkcolor=linkcol,%
            citecolor=citecol,%
            urlcolor=urlcol,%
            hypertexnames=true,%
            pdfpagelabels,
	    ]%
      {hyperref}

\usepackage[OT2,OT1]{fontenc}
\newcommand\cyr
{
\renewcommand\rmdefault{wncyr}
\renewcommand\sfdefault{wncyss}
\renewcommand\encodingdefault{OT2}
\normalfont
\selectfont
}
\DeclareTextFontCommand{\textcyr}{\cyr}

\usepackage[compact]{titlesec}
\titlespacing*{\section}{0pt}{*4}{4pt} 
\titlespacing{\subsection}{0pt}{*3}{3pt}

\definecolor{linkcol}{rgb}{0,0,0.5}
\definecolor{citecol}{rgb}{0,0.5,0.3}
\definecolor{urlcol}{rgb}{0.3,0,0}
\renewenvironment{thebibliography}[1]{
  \begin{oldthebibliography}{#1}
    \setlength{\itemsep}{0.1em}
    \setlength{\parskip}{0.1em}
}
{
  \end{oldthebibliography}
}

\renewcommand{\footnoterule}{%
  \kern -3pt
  \hrule width 1in
  \kern 2pt
}

\usepackage{xspace}

\makeatletter
\def\url@leostyle{%
  \@ifundefined{selectfont}{\def\UrlFont{}}%
  {\def\UrlFont{}}%
}
\makeatother
\usepackage[hyphenbreaks]{breakurl}

\newif\ifwatermark
\watermarkfalse

\ifwatermark
    \usepackage{draftwatermark}
    \SetWatermarkScale{.7}
    \SetWatermarkText{\shortstack{In Submission:\\Do Not Distribute}}
    \SetWatermarkColor[rgb]{1,0.88,0.88}
\fi

\usepackage{etoolbox}
\makeatletter
\patchcmd\@combinedblfloats{\box\@outputbox}{\unvbox\@outputbox}{}{%
   \errmessage{\noexpand\@combinedblfloats could not be patched}%
}%
 \makeatother
\AtBeginShipout{%
  \ifnum\value{page}>1 %
    \typeout{* Additional boxing of page `\thepage'}%
    \setbox\AtBeginShipoutBox=\hbox{\copy\AtBeginShipoutBox}%
  \fi
}

\usepackage{amsthm}
\theoremstyle{definition}
\newtheorem{exmp}{Example}

\begin{document}

\title{\LARGE \bf A Self-Attentive Emotion Recognition Network}

\author{Harris Partaourides$^{\star}$, Kostantinos Papadamou$^{\star}$, Nicolas Kourtellis$^+$, Ilias Leontiadis$^\ddagger$ \thanks{$^\ddagger$Work done while at Telefonica Research}, Sotirios Chatzis$^{\star}$\\[0.5ex]
\normalsize $^{\star}$Cyprus University of Technology, $^{+}$Telefonica Research, $^\ddagger$Samsung Research\\ 
\normalsize c.partaourides@cut.ac.cy, ck.papadamou@edu.cut.ac.cy, nicolas.kourtellis@telefonica.com, \\%
\normalsize i.leontiadis@samsung.com, sotirios.chatzis@cut.ac.cy}
\date{}

\maketitle
\begin{abstract}
Modern deep learning approaches have achieved groundbreaking performance in modeling and classifying sequential data. 
Specifically, attention networks constitute the state-of-the-art paradigm for capturing long temporal dynamics. 
This paper examines the efficacy of this paradigm in the challenging task of emotion recognition in dyadic conversations. 
In contrast to existing approaches, our work introduces a novel attention mechanism capable of inferring the immensity of the effect of each past utterance on the current speaker emotional state.
The proposed attention mechanism performs this inference procedure without the need of a decoder network; this is achieved by means of innovative self-attention arguments. 
Our self-attention networks capture the correlation patterns among consecutive encoder network states, thus allowing to robustly and effectively model temporal dynamics over arbitrary long temporal horizons.
Thus, we enable capturing strong affective patterns over the course of long discussions.
We exhibit the effectiveness of our approach considering the challenging IEMOCAP benchmark. 
As we show, our devised methodology outperforms state-of-the-art alternatives and commonly used approaches, giving rise to promising new research directions in the context of Online Social Network (OSN) analysis tasks. 
\end{abstract}

\section{Introduction}
Affective computing is an interdisciplinary research field that aims at bridging the gab between human and machine interactions.
To that end, researchers utilize sentiment analysis \cite{li2009non,ren2016improving,wilson2005recognizing} and emotion recognition \cite{alm2005emotions,kim2010evaluation,strapparava2008learning} algorithms to develop systems that can recognize emotions to properly drive their responses. 
In this context, the accuracy of emotion recognition is crucial for the success of affective computing solutions. 
Therefore, it is needed that the machine learning community develops increasing complex models, far and beyond the models used in the related but simpler task of sentiment analysis. 
Traditionally, to successfully recognize emotions, researchers have to utilize a variety of modalities such as speech, facial expression, body gestures and physiological indices \cite{holzman2003classification,kim2013deep,tripathi2017using}. 
This combination of distinct modalities ensures algorithm effectiveness.

This work is motivated from the important challenge of online emotion recognition from textual dialog data (online chats).
This is a problem of increasing immensity due to the emergence and strong popularity of Online Social Networks (OSNs).
Unfortunately, existing algorithms that address this problem suffer from two major shortcomings: 
1) they cannot capture temporal affective patterns over long dialogs; this results in missing crucial information that may have appeared many utterances before, but has salient effect on the current emotional state of the speakers, 
2) as they have access to only one type of modality, namely text, these algorithms cannot achieve high recognition performance since this typically requires combination of multiple modalities. 

In the affective computing literature, we observe a plethora of machine learning algorithms used for emotion recognition, such as linear discriminant classifiers, k-nearest neighbor, decision tree algorithms, support vector machines \cite{aman2008using,yang2007emotion} and artificial neural networks \cite{zadeh2017tensor}.
More recently, the research community has shown that machine learning models with the capacity to capture contextual information are capable of achieving much higher performance, as is well expected due to the nature of dialog data \cite{agrawal2012unsupervised,cauldwell2000did,ren2016context,vanzo2014context}.
Indeed, using contextual information to perform emotion recognition is similar to the actual process people use to infer the emotional state of their interlocutor. 
This becomes apparent in cases where the latest utterance is insufficient in inferring emotions. 
In such cases, people consider how the conversation evolved to acquire the missing information.
Hence, it is indisputable that, more often than not, we need contextual information to accurately predict emotional states from dialog text.

\begin{figure}[t!]
\centering
\includegraphics[width=\columnwidth]{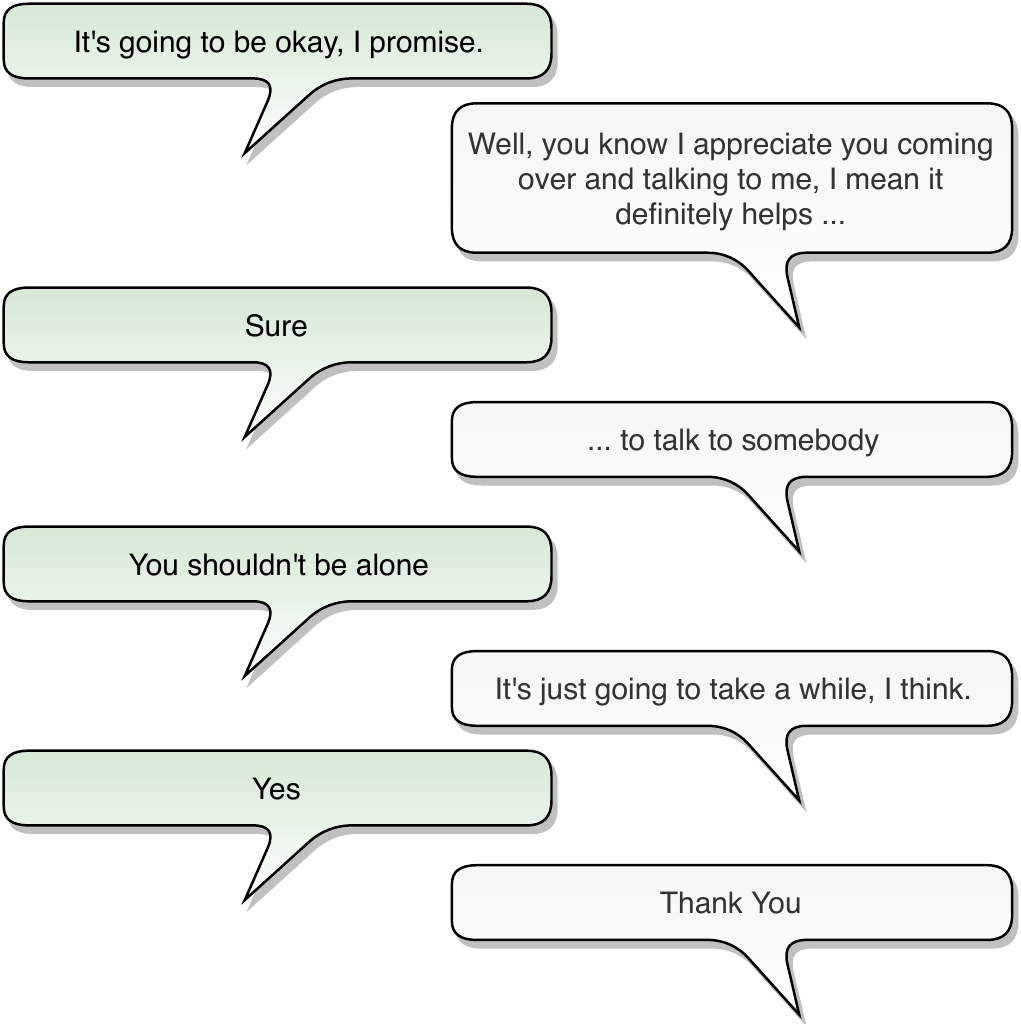}
\caption{A typical dialog segment (extracted from the IEMOCAP dataset \cite{busso2008iemocap})} 
\label{fig1}
\end{figure}

\begin{exmp}
In the conversation shown in Fig. \ref{fig1}, the last message (``Thank you'') does not imply anything about the speaker's emotional state. 
However, by analyzing the conversation up to that point, one can clearly infer the underlying emotional state (sadness).
\end{exmp}

To address these shortcomings, researchers have relied on models that can capture temporal dynamics, including hidden Markov models (HMMs) and recurrent neural networks (RNNs).
These context-aware models have yielded major improvements compared to their context-unaware counterparts \cite{poria2017context,wollmer2010context}.
However, both HMMs and RNNs suffer from a major limitation that undermines the effectiveness of emotion recognition in the context of OSNs dialogs: they both are model families that can capture temporal dependencies over short horizons. 
This implies a clear inability to retain salient information that may affect emotion over a long horizon, spanning the whole course of an OSN dialog.

Recently, the machine learning community has attempted to achieve a breakthrough in the performance of emotion recognition systems by relying on neural attention mechanisms \cite{hazarika2018conversational}.
These mechanisms build upon the short-term memory capabilities of RNNs to enable the creation of strong machine learning pipelines that can capture salient temporal dynamics over long horizons.
However, their efficacy has been examined only in the context of modeling fused distinct modalities, including speech and facial gestures, in addition to text. 
Besides, existing works \cite{khurana2018resolving,yang2016hierarchical} have relied on single-layer Bidirectional RNNs (BiRNNs) \cite{schuster1997bidirectional} for encoding context-level dialog dynamics; a fact that requires a-priori provision of the whole dialog to perform inference. 
This is clearly limiting, as it renders rather prohibitive the real-time analysis of OSN activity.

This paper offers a coherent solution that addresses the aforementioned limitations of the existing neural attention paradigm in the context of online emotion recognition from OSN dialog text. 
For the first time in the literature, we introduce a \emph{self-attentive hierarchical encoder} network that is capable of extracting salient information on both the individual utterance level as well as the level of the dialog context, as it has evolved until any given time point. 
Specifically, our proposed model comprises a hierarchical encoding mechanism that performs representation extraction on two levels: The first employs a Bidirectional Long Short Term Memory (LSTM) \cite{hochreiter1997long} that captures word-level contextual information in each individual utterance. 
The second utilizes a GRU \cite{cho2014learning} that performs dialog context-level representation to allow for capturing the salient dynamics over the whole dialog span. 

The formulated hierarchical encoder is complemented with a novel self-attention (SA) mechanism. 
This is carefully designed to generate accurate inferences of how strongly the encoder-obtained representations of the latest utterance at the final layer (\emph{dialog state}) correlate with the corresponding representations pertaining to previous utterances. 
As these representations constitute RNN states, which inherently encode short-term dynamics, the so-obtained corelation information allows for establishing a notion of attention among the current and the previous \emph{dialog states}.
Therefore, this self-attention information can be leveraged to yield meaningful weights for effectively combining the whole history of dialog states into a highly informative dialog context vector; we eventually use the resulting self-attentive context vector to drive an accurate penultimate emotion recognition layer of high accuracy. 
We emphasize that our use of a simple GRU at the second level of the encoder, as apposed to a bidirectional one, allows for performing emotion inference without requiring a-priori provision of the whole dialog, that is in an online fashion.
We dub our proposed approach the Self-attentive Emotion Recognition Network (SERN).

We experimentally evaluate our approach on the IEMOCAP dataset \cite{busso2008iemocap} and empirically demonstrate the significance of the introduced self-attention mechanism.
Subsequently, we perform an ablation study to demonstrate the robustness of the proposed model.
We empirically show an important enhancement of the attainable speaker emotional state inference capabilities. 
This is of vital importance for OSNs, since they are increasingly associated with distress and negative implications on users' mental health \cite{chen2013sharing}.

The remainder of this paper is organized as follows.
Section 2 provides a concise review of the methodological background of our approach.
In Section 3, we introduce the proposed approach and elaborate on its training and inference algorithms. 
Section 4 constitutes an in-depth experimental evaluation of the proposed method using a popular benchmark dataset. 
Finally, in Section 5 we summarize our contribution and conclude this paper by discussing directions for future research.

\section{Methodological Background}

\subsection{Word Representations}
In order for machine learning algorithms to perform analysis of word data, it is needed that the observed words are transformed into a vectorial representation; these are typically referred to as word embeddings in the related literature.
Word2Vec~\cite{mikolov2015computing} is a popular embedding technique based on deep learning principles. 
It aims at yielding embedding spaces of low dimensionality and high representational power.
This is achieved by postulating a one-hidden-layer softmax classifier which is presented with sentence fragments of fixed length, and is trained to predict the next word in the corresponding sentences.
In a different vein, GloVe~\cite{pennington2014glove} is an unsupervised algorithm for obtaining vector representations of words. 
Its main principle consists in capturing word-word concurrences based on the frequency that dictionary words co-occur in the available training corpus.
In this work, we relay on the Word2Vec scheme; however, we elect to train the representations from scratch, using our available datasets, as opposed to resorting to the pretrained Word2Vec embeddings. 

\subsection{Recurrent Neural Networks}
A recurrent neural network (RNN) is a neural network with the capacity to model the temporal dependencies in sequential data. 
Given an input sequence $x=(\mathbf{x}_1,...,\mathbf{x}_T)$, an RNN computes the hidden sequence $h=(\mathbf{h}_1,...,\mathbf{h}_T)$; its hidden vector $\mathbf{h}_t$ constitutes a concise representation of the temporal dynamics in a short-term horizon prior to time $t$. 
At each time step $t$, the hidden state $\mathbf{h}_t$ of the RNN is updated by $\mathbf{h}_t = f(\mathbf{h}_{t-1}, \mathbf{x}_t)$ where $f$ is a non-linear activation function. 
Given the state sequence $h$, the network eventually computes an output sequence $y=(\mathbf{y}_1,...,\mathbf{y}_T)$. 
A significant advantage of RNNs is the fact that they impose no limitations on input sequence length. 
However, practical application has shown that RNNs have difficulties in modeling long sequences. 
Specifically, RNNs are notorious for the exploding and vanishing gradients problem, which renders model training completely intractable for applications that entail long sequences. 

To resolve these issues, two popular RNN variants are usually employed, namely the GRU \cite{cho2014learning} and the LSTM \cite{hochreiter1997long} network.
The hidden state, $\mathbf{h}_t$, of a GRU network at time $t$ is given by:
\begin{equation}
\mathbf{z}_t = \sigma(W_z \mathbf{x}_t + U_z \mathbf{h}_{t-1} + \mathbf{b}_z)
\end{equation}
\begin{equation}
\mathbf{r}_t = \sigma(W_r \mathbf{x}_t + U_r \mathbf{h}_{t-1} + \mathbf{b}_r)
\end{equation}
\begin{equation}
\mathbf{h}_t = (1 - \mathbf{z}_t) \circ \mathbf{h}_{t-1} + \mathbf{z}_t \circ \tanh(W_h \mathbf{x}_t + U_h(\mathbf{r}_t \circ \mathbf{h}_{t-1}) + \mathbf{b}_h)
\end{equation}
On the other hand the hidden state, $\mathbf{h}_t$, of an LSTM network at time $t$ is given by:
\begin{equation}
\mathbf{f}_t = \sigma(W_f \mathbf{x}_t + U_f \mathbf{h}_{t-1} + \mathbf{b}_f)
\end{equation}
\begin{equation}
\mathbf{i}_t = \sigma(W_i \mathbf{x}_t + U_i \mathbf{h}_{t-1} + \mathbf{b}_i)
\end{equation}
\begin{equation}
\mathbf{o}_t = \sigma(W_o \mathbf{x}_t + U_o \mathbf{h}_{t-1} + \mathbf{b}_o)
\end{equation}
\begin{equation}
\mathbf{c}_t = \mathbf{f}_t \circ \mathbf{c}_{t-1} + \mathbf{i}_t \circ \tanh(W_c \mathbf{x}_t + U_c \mathbf{h}_{t-1} + \mathbf{b}_c)
\end{equation}
\begin{equation}
\mathbf{h}_t = \mathbf{o}_t \circ \tanh(\mathbf{c}_t)
\end{equation}
In these equations, the $W_.$, $U_.$, $\mathbf{b}_.$ are the trainable parameters and $\sigma$ is the logistic sigmoid function.

Finally, bidirectional formulations of RNNs have great use in natural language processing tasks.
Specifically, when dealing with understanding of whole sentences, it is intuitive to jointly model the temporal dynamics that stem from reading the sentence both in a forward and a backward fashion. 
Indeed, this may facilitate a more complete extraction of syntactic context, which is crucial for language understanding.
In essence, bidirectional RNN variants comprise two distinct RNNs, one presented with the observed sequence, and one presented with its reverse.
At each time point, the state vectors $\mathbf{h}_t$ of the two component RNNs are concatenated and presented as such to the penultimate layer of the network.

\subsection{Neural Attention}

Neural attention has been a major breakthrough in Deep Learning for Natural Language Processing, as it enables capturing long temporal dynamics that elude the capacity of RNN variants.
Among the large collection of recently devised neural attention mechanisms, the vast majority build upon the concept of soft attention \cite{xu2015show}. 
Given a sequence of hidden states $\mathbf{h}_t$ ($t=1,..,T$), the attention mechanism computes the context vectors, $\mathbf{c}_t$; these are  weighted sums of the available hidden states and are given by:
\begin{equation}
a^t_s = softmax(score(\mathbf{h}_t,\mathbf{h}_s)) ,\; t \neq s
\end{equation}
\begin{equation}
\mathbf{c}_t = \sum_s a^t_s \mathbf{h}_s
\end{equation}
In this expression, typical options for the score function are:
\begin{equation}
score(\mathbf{h}_t,\mathbf{h}_s) = 
\left \{
\begin{tabular}{c}
$\mathbf{h}^T_t \mathbf{h}_s$ \\
$\mathbf{h}^T_t W_a \mathbf{h}_s$ \\
$\mathbf{u}^T_a \tanh( W_a[\mathbf{h}_t;\mathbf{h}_s])$ 
\end{tabular}
\right.
\end{equation}
where the $W_a$ and $\mathbf{u}_a$ are trainable parameters.
In cases where we are dealing with a model generating whole sequences of different length from the input sequence, the $\mathbf{h}_s$ is the current hidden state of the sequence-generating model component, also known as the decoder.
On the other hand, when dealing with frame-level classification tasks where the penultimate network layer is a softmax classifier, as opposed to a decoder, the $\mathbf{h}_s$ can be the current state of the employed RNN, yielding 
\begin{equation}\label{eq:12}
a^t_s = softmax(score(\mathbf{h}_t,\mathbf{h}_s)) ,\; t \leq s
\end{equation}
\begin{equation}\label{eq:13}
\mathbf{c}_s = \sum_{t \leq s} a^t_s \mathbf{h}_t
\end{equation}

\begin{figure*}[t!]
  \centering
  \includegraphics[width=0.9\textwidth, height=7.5em]{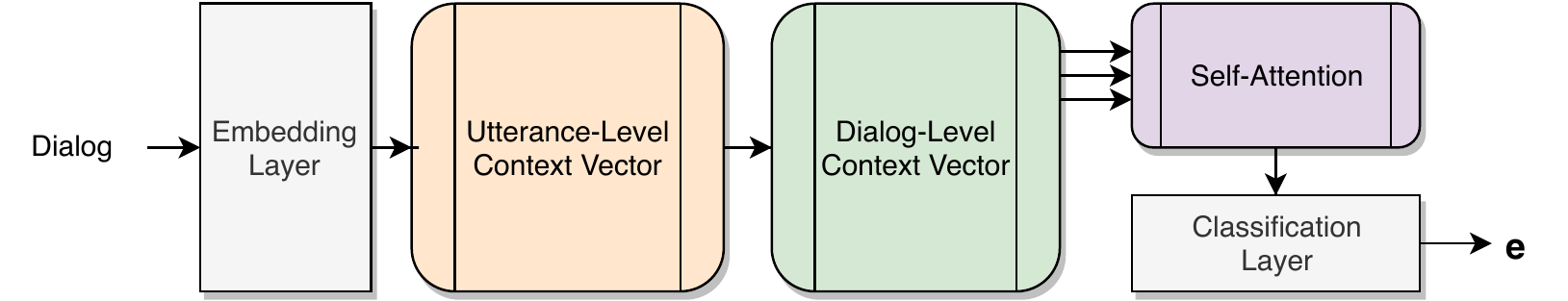}
  \caption{SERN model configuration.}
  \label{fig:DRNN-SA}
\end{figure*}

\section{Proposed Approach}

As already discussed, the ultimate goal of this work is to enable accurate emotion recognition in online text chats.
This gives rise to the challenging task of performing natural language understanding at both the utterance level and the dialog context up to the current utterance.
This is a problem of immense complexity, since it requires the capacity to perform valid inference at the utterance level and effectively correlate the obtained information over arbitrarily long dialog durations.

To this end, we introduce a novel hierarchical encoder network that is capable of extracting salient information on the individual utterance level, and inferring potent temporal dynamics across the dialog duration.
The latter capacity is enabled by appropriately implementing the concept of self-attention as an intrinsic part of our novel architecture. 

Let us consider a dialog $X = \{ \mathbf{X}_1, ..., \mathbf{X}_s\}$, where the $\mathbf{X}_s$ represents the $s$-th utterance of the dialog, and $\mathbf{X}_s = \{ X_{s_1}, ... X_{sw_s} \}$, where $w_{s}$ represents the word count in the $s$-th utterance.
We seek a model capable of correctly recognizing the dominant emotion related to each utterance, summarized into the vector $\mathbf{e} \in E^S$, where $E$ is the considered addressed set of emotions. 
In the following, we consider the six main emotion categories, namely angry, happy, sad, neutral, excited and frustrated; these are the emotions with an adequate amount of examples in the IEMOCAP dataset.
A descriptive illustration of the proposed model configuration is provided in Fig \ref{fig:DRNN-SA}. 

Based on this motivation, our proposed approach comprises three consecutive core parts: 
Initially, a trainable \emph{Word2Vec embeddings} mechanism is presented with the input sequence. 
Let us denote as $\mathbf{m}_{st}$ the word embedding pertaining to the $t$-th token of the $s$-th utterance.
Subsequently, a \emph{bidirectional LSTM} (BiLSTM) is used to capture the salient lingual information contained within each utterance. 
We use a bidirectional LSTM to ensure optimal inference of syntactic structure at the utterance level, as typical in the literature.
Let $\mathbf{f}_s^{utt}$ be the final state vector of the employed \emph{utterance-level BiLSTM}, presented with the $s$-th utterance.
This constitutes the latent vector representation fed to the \emph{subsequent dialog-level GRU} network.
Specifically, this network uses the BiLSTM-obtained latent vector representations of the preceding utterances to infer salient temporal dynamics at the dialog-context level, useful for driving a penultimate emotion classification layer.

Let us denote a running dialog comprising $s$ utterances. 
The postulated GRU network presented with the utterance-level representations \{ $\mathbf{f}_{s'}^{utt} \}_{s'=1}^s$ has generated a set of state vectors \{ $\mathbf{f}_{s'}^{dial} \}_{s'=1}^s$ representing dialog-level semantic information. 
This could be used to drive a penultimate dialog context-informed emotion classification layer.
However, as already discussed, RNN variants are only capable of capturing temporal dependencies over short-term horizons, with exponentially-decreasing temporal effect.
As real-world dialogs may be quite long and entail a gradual temporal evolution that spans long time frames, it is imperative that we endow the proposed model with the capacity to capture such long temporal dynamics. 

To this end, we deploy a \emph{self-attention layer} on top of the dialog-level GRU network.
As discussed in Section 2.3, the postulated self-attention mechanism computes, for the current utterance $s$, the affinity of its dialog-level representation, $\mathbf{f}_{s}^{dial}$, with the representations pertaining to the previous utterances.
On this basis, it computes an affinity-weighted average of these representations, as described in Eqs. \ref{eq:12}-\ref{eq:13}, which eventually drives emotion classification.
We refer to this weighted average as the dialog context vector at step $s$, $\mathbf{c}_s$.

We train the devised model in an end-to-end fashion. 
The employed training objective function is the categorical cross-entropy of the model; this is a natural selection, as we are dealing with a frame-level classification problem.
Specifically, we resort to stochastic gradient descent to obtain parameter estimators, employing the Adam optimizer \cite{kingma2014adam}.

\section{Experimental Evaluation}

In this section, we perform a thorough experimental evaluation of our proposed model. 
We provide a quantitative assessment of the efficacy and effectiveness of SERN, combined with deep qualitative insights pertaining to the functionality of the self-attention scheme.
Furthermore, we perform an ablation study to better illustrate the robustness of our approach.
To this end, we utilize a well-known benchmark for emotion recognition, namely the IEMOCAP dataset \cite{busso2008iemocap}. 
We have implemented our model in TensorFlow \cite{tensorflow2015-whitepaper}.
The code of our implementation can be found on GitHub\footnote{\url{https://github.com/Partaourides/SERN}}.

\subsection{IEMOCAP Dataset}

The Interactive Emotional Dyadic Motion Capture (IEMOCAP) database has been collected by emulating conversations in a controlled environment in order to study expressive human behaviors.
The conversations have been performed by ten unique speakers over five dyadic sessions in both a scripted but also an improvisation manner, with various audio-visual modalities being recorded. 
Each utterance in the dataset is labeled by three human annotators using categorical labels; these include angry, sad, happy, frustrated, excited, neutral, as well as other categories which we omit in this study. 
The available annotation has been performed by three annotators who assess the emotional states of the speakers taking into consideration dialog context. 
Thus, this dataset requires that we employ a model capable of inferring potent dialog-level contextual dynamics, as is the case with the proposed approach.

In our experiments, we only utilize the textual modality (transcripts) of the dataset and the categorical labels of each utterance. 
The used label information is derived by performing majority voting on the three available annotations.
This dataset contains $151$ conversations with a total number of $10,039$ utterances.
However, only $7,380$ utterances contain the six types of emotions we retain in this study; thus, the remaining utterances are omitted.
Table \ref{tab:iemocap_dataset_classes_details} provides a breakdown of the resulting dataset.

\begin{table}[t!]
\resizebox{\columnwidth}{!}{%
\begin{tabular}{lrrrrrr}
\hline
\textbf{Class} & Angry & Sad & Happy & Frustrated & Excited & Neutral \\ 
\hline
\textbf{Utterances} & 1,103 & 1,084 & 595 & 1,849 & 1,041 & 1,708 \\
\hline
\end{tabular}
}
\caption{IEMOCAP dataset: Number of utterances on each emotion.}
\label{tab:iemocap_dataset_classes_details}
\end{table}

Our data pre-processing regimen consists in word-based sentence segmentation and removing words with low frequency ($frequency<5$); to this end, we use the NLTK~\footnote{https://www.nltk.org/} library.
Similar with \cite{poria2017context}, we split the dataset into a training and test set by leaving out the fifth dyadic session. 
To perform hyperparameter tuning, we hold out a small representative subset of the training set ($\sim$ 7\%) to form a validation set.
Table~\ref{tab:iemocap_train_test_sets_details} summarizes the details of this split.

\begin{table}[t!]
\centering
\resizebox{\columnwidth}{!}{%
\begin{tabular}{lrrrrrrr}
\hline
\textbf{Class} & Angry & Sad & Happy & Frustrated & Excited & Neutral & Dialogs\\ 
\hline
\textbf{Train+Validation} & 933 & 839 & 452 & 1,468  & 742 & 1,324 & 120 \\
\textbf{Test} & 170 & 245 & 143 & 381 & 299 & 384 & 31 \\
\hline
\end{tabular}
}
\caption{Model training: Dataset split.}
\label{tab:iemocap_train_test_sets_details}
\end{table}

\subsection{Quantitative Study}

To exhibit the effectiveness of our approach, we compare its performance with the following baselines: 
\begin{itemize}
  \item \textbf{Support Vector Machine ($SVM$):} A simple classifier that does not consider utterance or dialog context information.
  \item \textbf{Bidirectional LSTM ($BiLSTM$):} A neural network with the capacity to capture only the utterance-level contextual information.
  \item \textbf{Bidirectional LSTM with self-attention ($BiLSTM_{att}$):} A \textbf{single-layer} $BiLSTM$ endowed with an additional self-attention mechanism, similar to Eqs. \ref{eq:12}-\ref{eq:13}.
\end{itemize}

In all cases, we perform stochastic gradient descent by means of the Adam algorithm with an initial learning rate of $5E^{-3}$, and epsilon, $1E^{-8}$.
Hyper-parameter tuning for the SVM is performed under the grid search strategy.
We train each model twenty times, with different initializations each time, and calculate the mean and standard deviation of the obtained accuracy, precision, recall and F1 scores.
We present our results in Table~\ref{tab:performance_metrics}.
For exhibition purposes, Table~\ref{tab:proposed_model_confusion_matrix} depicts the confusion matrix obtained from a randomly picked experiment repetition, combined with the corresponding precision and recall metrics.

\begin{table}[t!]
\resizebox{\columnwidth}{!}{%
\begin{tabular}{lcccc}
\hline
 & \textbf{Accuracy} & \textbf{Precision} & \textbf{Recall} & \textbf{F1 Score} \\ 
\hline
SVM & 0.313 ($\pm 0.00$) & 0.484 ($\pm 0.00$) & 0.235 ($\pm 0.00$) & 0.316 ($\pm 0.00$) \\ 
$BiLSTM$ & 0.477 ($\pm 0.01$) & 0.471 ($\pm 0.02$) & 0.459 ($\pm 0.01$) & 0.465 ($\pm 0.01$) \\ 
$BiLSTM_{att}$ & 0.516 ($\pm 0.02$) & 0.516 ($\pm 0.02$) & 0.501 ($\pm 0.02$) & 0.509 ($\pm 0.02$) \\ 
SERN & 0.522 ($\pm 0.02$) & 0.544 ($\pm 0.02$) & 0.517 ($\pm 0.02$) & 0.530 ($\pm 0.02$) \\ 
\hline
\end{tabular}
}
\caption{Performance metrics.}
\label{tab:performance_metrics}
\end{table}

\begin{table}[t!]
\resizebox{\columnwidth}{!}{%
\begin{tabular}{lrrrrrr|r}
\hline
& \multicolumn{1}{r}{\textbf{Angry}} & \multicolumn{1}{r}{\textbf{Excited}} & \multicolumn{1}{r}{\textbf{Frustrated}} & \multicolumn{1}{r}{\textbf{Happy}} & \multicolumn{1}{r}{\textbf{Neutral}} & \multicolumn{1}{r|}{\textbf{Sad}} & \multicolumn{1}{r}{\textbf{Recall}} \\ 
\hline
\textbf{Angry} & 110 & 2 & 29 & 0 & 22 & 7 & 0.647 \\ 
\textbf{Excited} & 9 & 156 & 8 & 74 & 27 & 25 & 0.522 \\ 
\textbf{Frustrated} & 71 & 6 & 193 & 1 & 87 & 23 & 0.507 \\ 
\textbf{Happy} & 14 & 19 & 0 & 80 & 29 & 1 & 0.559 \\ 
\textbf{Neutral} & 35 & 34 & 83 & 11 & 197 & 24 & 0.513 \\
\textbf{Sad} & 9 & 12 & 42 & 7 & 11 & 164 & 0.669 \\ 
\hline
\textbf{Precision} & 0.444 & 0.681 & 0.544 & 0.462 & 0.528 & 0.672 & \\
\hline
\end{tabular}%
}
\caption{A randomly-picked confusion matrix and the corresponding precision and recall metrics. The confusion matrix rows and columns depict the ground-truth and the predicted emotions, respectively.}
\label{tab:proposed_model_confusion_matrix}
\end{table}

As we observe, SERN yields notable performance improvements over the alternatives in all performance metrics.
We emphasize that application of the Student's t-test corroborates the statistical significance of the observed differences among the alternatives.
Specifically, the p-values obtained on all performance metrics is below the 0.05 threshold.
The confusion matrix of Table~\ref{tab:proposed_model_confusion_matrix} depicts the number of accurate and misclassified predictions; this illustrates the difficulties in predicting the actual emotional state of the speaker.
Apparently, the most prominent difficulties arise between the (neutral, frustrated) and (excited,happy) pairs. 

\begin{table}[t!]
\includegraphics[width=\linewidth]{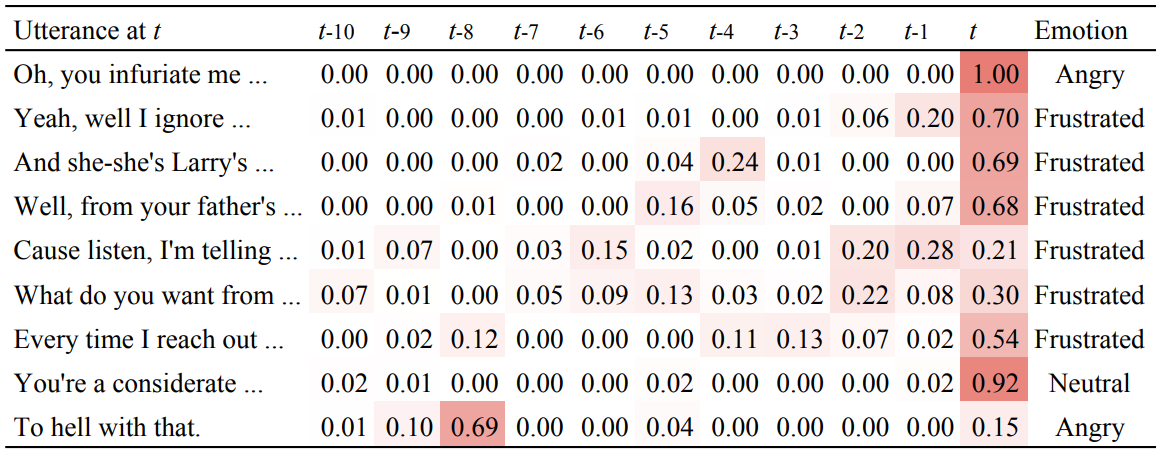}
\caption{Example A. Self-attention weights over windows spanning the latest nine utterances. $t$ denotes the current timestep in the dialog.}
\label{tab:attention_weights_dialog_example_a}
\end{table}

\begin{table}[t!]
\includegraphics[width=\linewidth]{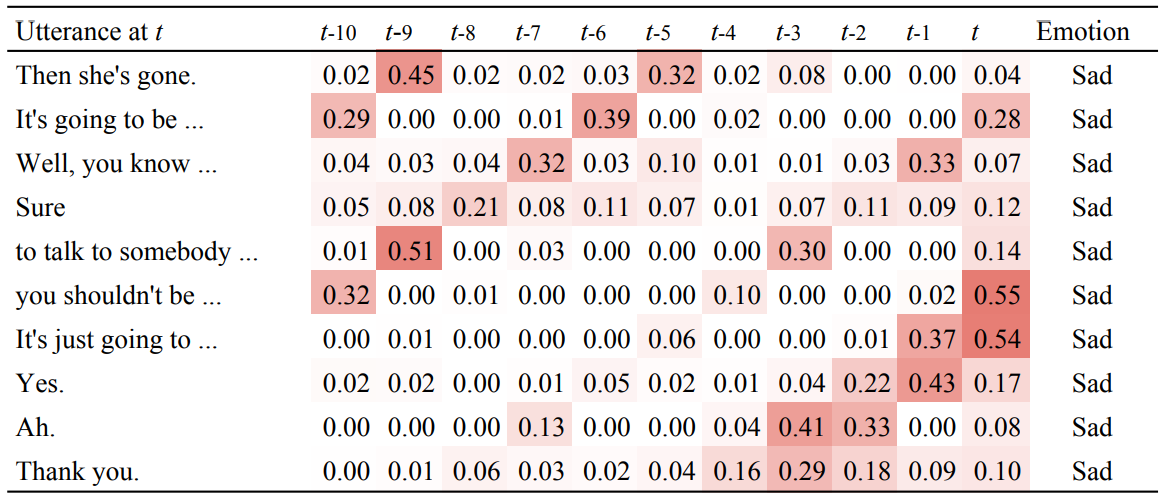}
\caption{Example B. Self-attention weights over windows spanning the latest ten utterances. $t$ denotes the current timestep in the dialog.}
\label{tab:attention_weights_dialog_example_b}
\end{table}

\subsection{Qualitative Study}

Here, we exhibit the role of the self-attention mechanism in enhancing the obtained emotion recognition accuracy.
To this end, we train a variant of SERN whereby the context vectors are computed over time-windows spanning the latest ten utterances, as opposed to whole course of the dialog.
In Tables \ref{tab:attention_weights_dialog_example_a} and \ref{tab:attention_weights_dialog_example_b}, we present the so-obtained self-attention weights over two representative dialog segments.
It becomes apparent that utterances many steps in the past may play a crucial role in describing the current emotional state; this salient information would have been missed had it not been for the employed self-attention mechanism.

For instance, the last message of the dialog segment shown in Table \ref{tab:attention_weights_dialog_example_b}, "Thank you," was uttered by a sad individual; this emotion can only be inferred through the utterances "you shouldn't be alone" and "It's just going to take a while, I think."
Alternatively, the last message of the dialog segment in Table \ref{tab:attention_weights_dialog_example_a}, "To hell with that," was uttered by an angry individual; this can be traced back to his/her emotional state during the first sentence, "Oh, you infuriate me sometimes. You know, it's not just my business if dad throws a fit."
Even more crucially, we observe that the last utterance is not often the principal component that drives the emotional state of the speaker. 
This can be observed at column $t$ of Tables \ref{tab:attention_weights_dialog_example_a} and \ref{tab:attention_weights_dialog_example_b} which contains self-attention weights which are less than the values at previous columns.

\subsection{Ablation Study}

To further assess the robustness of our model, we consider a different mixture of recognized emotions. 
First, we train and test it with five emotions (angry, happiness, sad, excited, neutral), hence ignoring the examples of the "frustrated" class; this is similar to \cite{busso2008iemocap}.
Then, we also train and test our model with four emotions (angry, happiness, sad, neutral), by merging the "excitement" and "happiness" categories to a single "happiness" category, similar to~\cite{lee2011emotion,rozgic2012ensemble}. 
In Table~\ref{tab:experiment_different_classes_models_preformance_metrics}, we present the obtained performance metrics, while in Table \ref{tab:experiment2_precision_per_class} we offer a breakdown for each emotion. 
To provide deeper insights, Tables \ref{tab:proposed_model_confusion_matrix_5} and \ref{tab:proposed_model_confusion_matrix_4} depict the confusion matrices obtained on a randomly-picked experiment repetition, combined with the corresponding precision and recall metrics.
We clearly observe that our method retains its robustness in these alternative settings.
Interestingly, the angry and neutral emotions become easier to discern when we omit the frustrated class (five classes scenario); a similar improvement is obtained when we combine the happy and excited emotions (four classes scenario).
We posit that this improvement stems from the high level of ambiguity between emotional types, even among human annotators. 

Finally, we examine the effect of the number of previous utterances used to compute the inferred context vectors.
To this end, we repeat our experiments with the SERN model using only the five, ten, twenty and fourty latest utterances to compute the context vectors, and compare to the outcome of the full-fledged model.
Our obtained results are depicted in Table \ref{tab:models_preformance_metrics}; it is obvious that using a window twenty steps long yields the best performance.

\begin{table}[t!]
\centering
\resizebox{\columnwidth}{!}{%
\begin{tabular}{ccccc}
\hline
\textbf{Classes} & \textbf{Accuracy} & \textbf{Precision} & \textbf{Recall} & \textbf{F1 Score} \\ 
\hline
4 & 0.689 ($\pm 0.03$) & 0.685 ($\pm 0.02$) & 0.699 ($\pm 0.02$) & 0.692 ($\pm 0.02$) \\ 
5 & 0.583 ($\pm 0.02$) & 0.589 ($\pm 0.02$) & 0.569 ($\pm 0.02$) & 0.579 ($\pm 0.02$) \\ 
6 & 0.522 ($\pm 0.02$) & 0.544 ($\pm 0.02$) & 0.517 ($\pm 0.02$) & 0.530 ($\pm 0.02$) \\ 
\hline
\end{tabular}
}
\caption{Performance metrics of $SERN$ trained on four, five and six emotions.}
\label{tab:experiment_different_classes_models_preformance_metrics}
\end{table}

\begin{table}[t!]
\resizebox{\columnwidth}{!}{%
\begin{tabular}{cccccccc}
\hline
\multicolumn{1}{r}{\textbf{Classes}} & \multicolumn{1}{r}{\textbf{Angry}} & \multicolumn{1}{r}{\textbf{Excited}} & \multicolumn{1}{r}{\textbf{Frustrated}} & \multicolumn{1}{r}{\textbf{Happy}} & \multicolumn{1}{r}{\textbf{Neutral}} & \multicolumn{1}{r}{\textbf{Sad}} & \multicolumn{1}{r}{\textbf{Happy+Excited}} \\ 
\hline
\textbf{4} & 0.617 & - & - & - & 0.720 & 0.667 & 0.847 \\ 
\textbf{5} & 0.649 & 0.767 & - & 0.317 & 0.685 & 0.635 & - \\ 
\textbf{6} & 0.444 & 0.681 & 0.544 & 0.462 & 0.528 & 0.672 & - \\
\hline
\end{tabular}
}
\caption{Classification precision breakdown for four, five and six detected emotions.}
\label{tab:experiment2_precision_per_class}
\end{table}

\begin{table}[t!]
\resizebox{\columnwidth}{!}{%
\begin{tabular}{lrrrrr|r}
\hline
& \multicolumn{1}{r}{\textbf{Angry}} & \multicolumn{1}{r}{\textbf{Excited}} & \multicolumn{1}{r}{\textbf{Happy}} & \multicolumn{1}{r}{\textbf{Neutral}} & \multicolumn{1}{r|}{\textbf{Sad}} & \multicolumn{1}{r}{\textbf{Recall}} \\ 
\hline
\textbf{Angry} & 122 & 0 & 0 & 35 & 13 & 0.718 \\ 
\textbf{Excited} & 7 & 102 & 121 & 36 & 33 & 0.341 \\ 
\textbf{Happy} & 10 & 18 & 71 & 27 & 17 & 0.497 \\ 
\textbf{Neutral}  & 34 & 8 & 21 & 278 & 43 & 0.724 \\ 
\textbf{Sad} & 15 & 5 & 11 & 30 & 184 & 0.751 \\ 
\hline
\textbf{Precision} & 0.649 & 0.767 & 0.317 & 0.685 & 0.635 & \\
\hline
\end{tabular}%
}
\caption{Five classes scenario: A randomly-picked confusion matrix and the corresponding precision and recall metrics.}
\label{tab:proposed_model_confusion_matrix_5}
\end{table}

\begin{table}[t!]
\resizebox{\columnwidth}{!}{%
\begin{tabular}{lrrrr|r}
\hline
& \multicolumn{1}{r}{\textbf{Angry}} & \multicolumn{1}{r}{\textbf{Happy+Excited}} & \multicolumn{1}{r}{\textbf{Neutral}} & \multicolumn{1}{r|}{\textbf{Sad}} & \multicolumn{1}{r}{\textbf{Recall}} \\ 
\hline
\textbf{Angry} & 116 & 1 & 37 & 16 & 0.682 \\ 
\textbf{Happy+Excited} & 19 & 337 & 42 & 44 & 0.762 \\ 
\textbf{Neutral} & 37 & 40 & 277 & 30 & 0.721 \\
\textbf{Sad} & 16 & 20 & 29 & 180 & 0.735 \\ 
\hline
\textbf{Precision} & 0.617 & 0.847 & 0.720 & 0.667 & \\
\hline
\end{tabular}%
}
\caption{Four classes scenario: A randomly-picked confusion matrix and the corresponding precision and recall metrics.}
\label{tab:proposed_model_confusion_matrix_4}
\end{table}

\begin{table}[t!]
\resizebox{\columnwidth}{!}{%
\begin{tabular}{lrrrr}
\hline
& \textbf{Accuracy} & \textbf{Precision} & \textbf{Recall} & \textbf{F1 Score} \\ 
\hline
$SERN_{5}$ & 0.557 & 0.563 & 0.552 & 0.558 \\ 
$SERN_{10}$ & 0.570 & 0.570 & 0.591 & 0.581 \\ 
$SERN_{20}$ & 0.584 & 0.583 & 0.580 & 0.582 \\ 
$SERN_{40}$ & 0.581 & 0.595 & 0.565 & 0.579 \\ 
$SERN$ & 0.555 & 0.555 & 0.570 & 0.562 \\ 
\hline
\end{tabular}
}
\caption{Performance metrics of the proposed model using different window size.}
\label{tab:models_preformance_metrics}
\end{table}

\section{Conclusions}

Accurate emotion recognition is a significant challenge for the developers and administrators of modern OSNs.
Indeed, the importance of these algorithms lies at facilitating the accurate and timely recognition of the emotional state of the speaker.
This renders them the key mechanism that could enable the development of effective mitigation strategies, for instance for dealing with cyberbullying and suicidal ideation in OSNs. 
However, this necessitates the availability of algorithms with high recognition accuracy.

In response to this need, in this paper we devised a self-attentive emotion recognition network that is composed of novel mixture of hierarchical encoding components and self-attention mechanisms.
Our overarching goal was to enable a more potent modeling of the dialog dynamics, with special emphasis on accounting for long-term affective inference.
Our formulation is carefully crafted to allow for predicting the emotional state of the speaker via a feed-forward scheme driven from the dialog evolution up to any desired time point.
This endows SERN with real-time capability, thus permitting its usage directly on OSNs.

We performed a thorough experimental evaluation of our approach using the challenging IEMOCAP benchmark.
We provided deep qualitative and quantitative insights to illustrate the efficacy of our modelling selections and the functional characteristics of our approach.
In addition, we performed comparisons to a number of state-of-the-art alternatives and showcased the superiority of our approach. 
These findings vouched for the usefulness of the introduced novel modeling arguments that underlie SERN. 

The promising findings of this work encourage us to pursue the further evolution of SERN.
We consider methodological extensions, for instance by exploring feedback-driven refinement by means of reinforcement learning techniques \cite{sutton2018reinforcement}. 
We also actively work on the integration of our model in a real-world OSN enhancement framework. 
These endeavors constitute our future research directives. 

\subsection*{Acknowledgments} 
This project has received funding from the European Union's Horizon 2020 Research and Innovation program under the Marie Sk\l{}odowska-Curie ENCASE project (Grant Agreement No. 691025).
This work reflects only the authors' views; the Agency and the Commission are not responsible for any use that may be made of the information it contains.

\small
\bibliography{bibliography}
\bibliographystyle{abbrv}

\end{document}